# Graph Neural Network Based Surrogate Model of Physics Simulations for Geometry Design

Jian Cheng Wong
*Department of Fluid Dynamics*
*Institute of High Performance Computing, A\*STAR*
Singapore
wongj@ihpc.a-star.edu.sg

Chin Chun Ooi
*Department of Fluid Dynamics*
*Institute of High Performance Computing & Center for Frontier AI Research, A\*STAR*
Singapore
ooicc@ihpc.a-star.edu.sg

Joyjit Chattoraj
*Department of Computing & Intelligence*
*Institute of High Performance Computing, A\*STAR*
Singapore
joyjit_chattoraj@ihpc.a-star.edu.sg

Lucas Lestandi
*Institut de Recherche en Génie Civil et Mécanique*
*Nantes Université, École Centrale Nantes, CNRS*
Nantes, France
lucas.lestandi@ec-nantes.fr

Guoying Dong
*Department of Mechanical Engineering*
*University of Colorado Denver*
Denver, USA
guoying.dong@ucdenver.edu

Umesh Kizhakkinan
*Digital Manufacturing and Design Center*
*Singapore University of Technology and Design*
Singapore
umesh_kizhakkinan@sutd.edu.sg

David William Rosen
*Digital Manufacturing and Design Center*
*Singapore University of Technology and Design*
Singapore
david_rosen@sutd.edu.sg

Mark Hyunpong Jhon
*Department of Engineering Mechanics*
*Institute of High Performance Computing, A\*STAR*
Singapore
jhonmh@ihpc.a-star.edu.sg

My Ha Dao
*Department of Fluid Dynamics*
*Institute of High Performance Computing, A\*STAR*
Singapore
daomh@ihpc.a-star.edu.sg

*Abstract*— Computational Intelligence (CI) techniques have shown great potential as a surrogate model of expensive physics simulation, with demonstrated ability to make fast predictions, albeit at the expense of accuracy in some cases. For many scientific and engineering problems involving geometrical design, it is desirable for the surrogate models to precisely describe the change in geometry and predict the consequences. In that context, we develop graph neural networks (GNNs) as fast surrogate models for physics simulation, which allow us to directly train the models on 2/3D geometry designs that are represented by an unstructured mesh or point cloud, without the need for any explicit or hand-crafted parameterization. We utilize an encoder-processor-decoder-type architecture which can flexibly make prediction at both node level and graph level. The performance of our proposed GNN-based surrogate model is demonstrated on 2 example applications: feature designs in the domain of additive engineering and airfoil design in the domain of aerodynamics. The models show good accuracy in their predictions on a separate set of test geometries after training, with almost instant prediction speeds, as compared to O(hour) for the high-fidelity simulations required otherwise.

## I. INTRODUCTION

Computational simulations of physics have become an essential component of modern science and engineering, and the field is growing rapidly. Physics simulations are frequently used as a model of reality to provide insights for understanding and predicting the behavior of physical systems at various levels of detail and accuracy. Simulation models can be computationally very expensive, due to their complex, multi-physics, and multi-scale nature. Therefore, sole use of high-fidelity simulation models for many tasks—such as material discovery, drug discovery, digital twin, computer-aided design—where one needs to quickly survey a vast number of possible scenarios and/or geometries, can quickly become intractable.

CI techniques have shown great potential to learn the nonlinear input-output relationship from existing simulation results, allowing for fast predictions, albeit at the expense of some accuracy. They are used as fast surrogate models of physics simulation to support discovery, design optimization, and decision-making process. Traditional surrogate models typically only predict a single output or a few outputs. The recent quest is to make predictions of the entire finely resolved physical field using advanced deep learning techniques. Deep convolutional neural networks, such as U-Net, have become a popular choice for the high dimensional surrogate modelling task when the data can be easily represented as 2D pixel or 3D voxel [1], and have been successfully applied to engineering problems such as flow past an airfoil [2]–[5], micro fluids channel flow [6], stress distribution in a 3D printed part [7], [8], and topological optimization [9].

Many scientific and engineering problems involve geometry design. It is desirable for the geometry, e.g., from molecular structure to the shape of airfoil, to be precisely represented by the surrogate models. The voxel-type representation however has its limitation when describing geometry designs with curvature and/or fine features, and hence this may impede the performance and applicability of convolutional neural network-based surrogate models. Usually, data augmentation will be required when training convolutional neural network models, e.g., to incorporate invariances. On the other hand, graph-based methods represent a promising candidate for modelling geometry [10], [11]. For example, in molecular and materials science domain, GNNs have been explored for learning the topology of atomic and molecular structures and predicting their properties [12], [13]. GNNs have also shown promise in predicting the time evolution of solid and fluid dynamic systems [14].

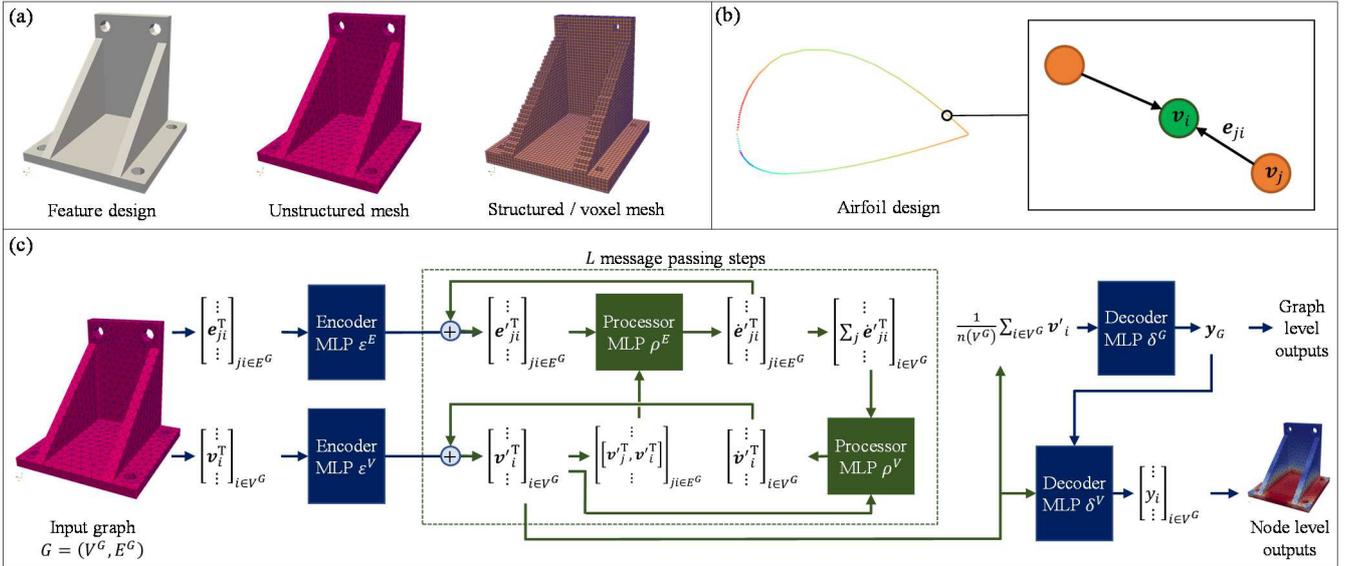

Figure 1: (a) Comparison of unstructured and structured / voxel meshes on an example geometry from the feature design problem (Section IV-A-1). The voxels are unable to describe features such as slope and hole, despite having a much denser mesh. (b) Example geometry from the airfoil design problem (Section IV-A-2) with nodes and edges. (c) Illustrative schematic of the GNN architecture employed.

In that context, we develop GNN-based surrogate models of physics simulation for geometry design. By utilizing graph representation, the model can more precisely describe the change in geometry with complex features and predict the consequences, which is essential for the design and optimization task. We evaluate the GNN-based surrogate models on 2 engineering design problems, i.e., feature design in the additive manufacturing domain and airfoil design in the aerodynamics domain. The models are not only capable of making fast prediction of an overall performance indicator, but also of predicting the entire physical field of a geometry design, such as residual stress distribution of a 3D printed part and surface pressure distribution on an airfoil, with orders of magnitude speed-up relative to typical high-fidelity simulations.

## II. STATEMENT OF THE PROBLEM

We consider the following surrogate modelling task. The goal is to develop a CI model that allows the computationally efficient evaluation of the physical quantities of interest, e.g., the physical field or an aggregated performance indicator, due to the change in geometry, in addition to any possible change in operating condition. The CI model is therefore required to map the input geometry and operating condition onto the output physical quantities of interest.

The data from physics simulations (e.g., finite element, finite volume, or finite difference methods) are used for training the CI model. Typically, simulations numerically compute the physical quantities on a *mesh*, i.e., a network of discretized cells and node points covering the continuous problem domain. In a nutshell, the simulation mesh can be classified into structured and unstructured mesh types. The structured mesh has a regular connectivity on homogenous, voxel-like cells. On the other hand, the unstructured mesh allows for mixed types of mesh cells and irregular connectivity, making it a preferred choice for handling complex geometries. Moreover, the amount of node points and connecting edges may change significantly given different geometry designs of varying shape, feature, and size. The CI model should be generic enough to learn from such unstructured mesh simulation data.

## III. GRAPH NEURAL NETWORK BASED SURROGATE MODEL OF PHYSICS SIMULATION

### A. Graph representation of simulation data

The unstructured simulation mesh with node points $V$ connected by mesh edges $E$ can be represented by a graph $G = (V^G, E^G)$. Each node $i \in V^G$ is associated with *node feature* $v_i$. Each edge $ij \in E^G$ connecting node $i$ to node $j$ is associated with *edge feature* $e_{ij}$. The graph contains bidirectional mesh edges, i.e., for each edge $ij \in E^G$ there exist an edge $ji \in E^G$ connecting from node $j$ to node $i$. Each node point is associated with a positional coordinate $x_i$ in Cartesian coordinate system, which is to be encoded into both node and edge features. To achieve spatial equivariance, we encode relative coordinate $x_i - x_{ref(G)}$ to a reference point $x_{ref(G)}$ in the given graph $G$ into the node feature $v_i$. We encode relative displacement vector $x_{ji} = x_j - x_i$ and its L2 norm $\|x_{ji}\|_2$ into the edge feature $e_{ji}$. The graph is a very general representation in the way that we may encode many other problem related parameters, such as the change in operating condition, into the node or edge features.

### B. Graph neural networks for learning simulation data

For this work, we use a GNN to learn the mapping from the input graph $G = (V^G, E^G)$ onto the output physical quantities that we want to model. We adopt an encode-process-decode type architecture (Figure 1), in accordance with prior work in literature [10], [14], [15].

**Encoder.** The encoder embeds each node and edge in the input graph into a latent vector of size $n_l$, using the encoder MLPs $\varepsilon^E$ and $\varepsilon^V$ for edge feature $e_{ji}$ and node feature $v_i$ respectively, i.e.,

$$e'_{ji} \leftarrow \varepsilon^E(e_{ji}), \quad (1a)$$

**Processor.** The processor recursively updates the embedded latent vectors at each node and edge with $L$ message passing steps. Each message passing step uses a separate block of processor MLPs $\rho^E$ and $\rho^V$ for the latent edge and node features:

$$\dot{e}'_{ji} \leftarrow \rho^E(e'_{ji}, v'_j, v'_i), \quad (2a)$$

$$\dot{v}'_i \leftarrow \rho^V(v'_i, \sum_j \dot{e}'_{ji}). \quad (2b)$$

The edge processor MLP $\rho^E$ first computes the update for latent edge features by taking the concatenated features of edge and its receiver and sender nodes from the previous step as input. Then, the node processor MLP $\rho^V$ takes the concatenated features of node from previous step and all incoming edges (with a sum pooling) at current step as input. Finally, the updates are added back to the latent edge and node features from previous step, creating a residual connection:

$$e'_{ji} \leftarrow e'_{ji} + \dot{e}'_{ji}, \quad (3a)$$

$$v'_i \leftarrow v'_i + \dot{v}'_i. \quad (3b)$$

**Decoder.** The latent node features are then passed to 2 separate decoders, i.e., a graph decoder MLP $\delta^G$ and a node decoder MLP $\delta^V$:

$$y_G \leftarrow \delta^G\left(\frac{1}{n(V^G)}\sum_{i \in V^G} v'_i\right), \quad (4a)$$

$$y_i \leftarrow \delta^V(v'_i, y_G). \quad (4b)$$

The graph decoder MLP extracts the graph level feature $y_G$ from the average pooling of all graph nodes. We denote $n(V^G)$ as the total number of nodes in $G$. For a graph level prediction task, $y_G$ becomes the output. For node level prediction task, $y_G$ will then be concatenated with the latent node features $v'_i$ and passed to the node decoder MLP $\delta^V$ to predict the physical quantities $y_i$ that we want to model.

## IV. RESULTS AND DISCUSSION

We evaluate the GNN-based surrogate models on 2 engineering problems, namely the feature design in the additive manufacturing domain and airfoil design in the aerodynamics domain.

### A. Description of problem domain and input encoding

#### 1) Feature design in additive manufacturing

The laser powder bed fusion process (LPBF) in additive manufacturing can fabricate parts with intricate geometries, providing a high level of design freedom for designers to improve the functional performance of products. While most existing studies in additive manufacturing domain have focused on simple geometry, e.g., regular, bulky shapes, such surrogate models may not be sufficient for the co-design of product and process. A more general surrogate model that can quickly evaluate the functional performance due to any change in shapes and features is needed. With the aim of predicting a wide range of geometries that can arise during the design process, [7] trained a U-Net model on a large set of 3D geometries comprising different combinations of basic geometric features, e.g., circular struts, square struts, and walls. The U-Net model however can have poor voxelization when representing certain features such as circular and thin struts.

In this work, we train a GNN on a similar set of geometries. The objective of the surrogate model is to predict the physical field, e.g., residual stress (measured through the scalar von Mises stress) on every node point, for features with different shape, sizes, orientations, and intersections. We utilize graph (unstructured mesh) representation which can more precisely describe these features (Figure 1a). Moreover, we do not need to perform the data augmentation as described in [7], since our encoding already accounts for spatial equivariance. 12 geometries (test set 1) are reserved for testing. The remaining 458 geometries (train set) are used for model training. Furthermore, 7 challenging geometries (test set 2) are designed to evaluate the out-of-distribution performance of the model: 3 contained more complex combinations of the circular struts, square struts, and walls; 2 are the original and optimized designs of an engineering bracket; the other 2 are the original and optimized designs of a bike stem. Selected feature designs are displayed in Figure 2. A high fidelity LPBF process full order simulation [16] was used to generate the ground truth for all these geometries.

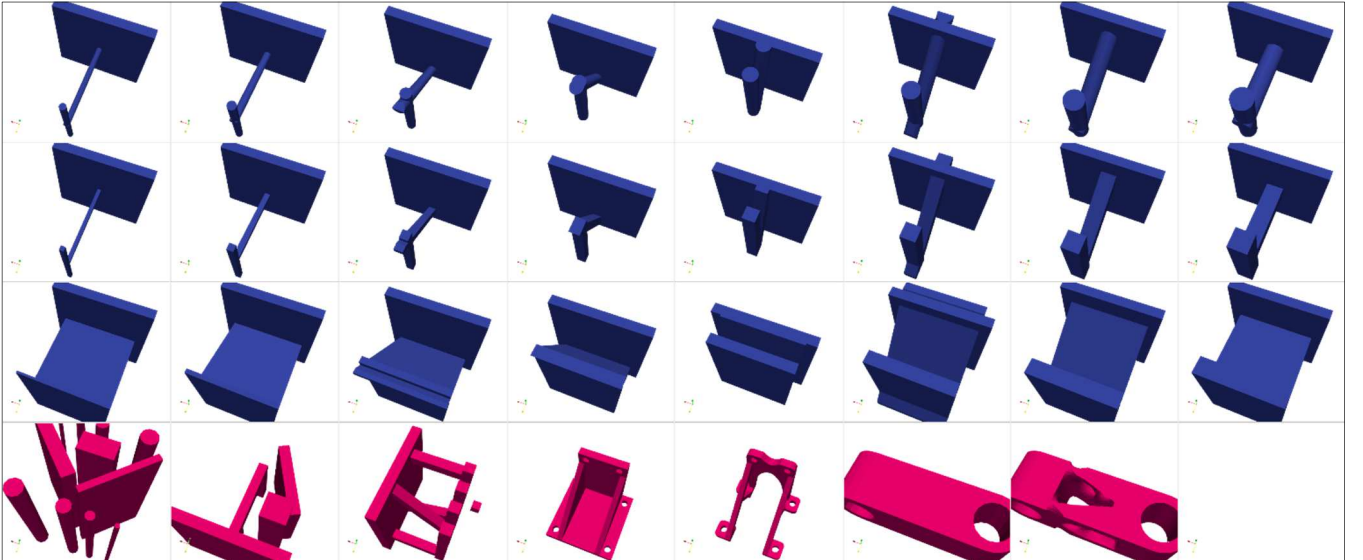

Figure 2: In blue: selected feature designs used for model training (train set). In red: 7 challenging geometries (test set 2) are designed to evaluate the out-of-distribution performance of the model.

TABLE I: Dataset, input encoding, architecture and training configurations used for GNN modelling in Section IV.

| Problem domain | No. training sample (graph) | Target output | Output type | Input graph: node $v_i$ | Input graph: edge $e_{ji}$ | MLP architecture (depth $n_d$, width $n_w$, output size $n_l$) for encoder $\varepsilon^E$, $\varepsilon^V$, processor $\rho^E$, $\rho^V$, and decoder $\delta^G$, $\delta^V$ | No. message passing step, $L$ | Training (loss, no. epoch, batch size, initial learning rate) |
|---|---|---|---|---|---|---|---|---|
| 1. Feature design (3D) | 458 ↑ 46 – 3492 nodes | Residual stress | Node level | $x_i - x_{ref(G)}$, $\|x_i - x_{ref(G)}\|_1$ $I_{cell\_type_i}$, $n_{e_i}$ | $x_{ji}$, $\|x_{ji}\|_2$ | $\varepsilon^E$, $\varepsilon^V$, $\rho^E$, $\rho^V$: $n_d$=4, $n_w$=64, $n_l$=64 $\delta^G$: $n_d$=4, $n_w$=64, $n_l$=4 $\delta^V$: $n_d$=4, $n_w$=64, $n_l$=1 | 6 | MAE + L1 reg. epoch=2000, batch_size=16, initial_lr=5e-4 |
| Airfoil design (2D) | 6372 ↑ 194 – 660 nodes | Pressure | Node level | $x_i - x_{ref(G)}$, $I_{u/l_i}$, $u_0, v_0$ | $x_{ji}$, $\|x_{ji}\|_2$ | $\varepsilon^E$, $\varepsilon^V$, $\rho^E$, $\rho^V$: $n_d$=5, $n_w$=64, $n_l$=64 $\delta^G$: $n_d$=5, $n_w$=64, $n_l$=4 $\delta^V$: $n_d$=5, $n_w$=64, $n_l$=1 | 5 | MAE + L1 reg. epoch=3000, batch_size=32, initial_lr=5e-4 |
| Airfoil design (2D) | 6372 ↑ 194 – 660 nodes | Drag coefficients, $C_D$ | Graph level | $x_i - x_{ref(G)}$, $I_{u/l_i}$, $u_0, v_0$ | $x_{ji}$, $\|x_{ji}\|_2$ | $\varepsilon^E$, $\varepsilon^V$, $\rho^E$, $\rho^V$: $n_d$=5, $n_w$=64, $n_l$=64 $\delta^G$: $n_d$=5, $n_w$=64, $n_l$=1 | 5 | MAE + L1 reg. epoch=500, batch_size=64, initial_lr=5e-4 |
| Airfoil design (2D) | 6372 ↑ 194 – 660 nodes | Lift coefficients, $C_L$ | Graph level | $x_i - x_{ref(G)}$, $I_{u/l_i}$, $u_0, v_0$ | $x_{ji}$, $\|x_{ji}\|_2$ | $\varepsilon^E$, $\varepsilon^V$, $\rho^E$, $\rho^V$: $n_d$=5, $n_w$=32, $n_l$=32 $\delta^G$: $n_d$=5, $n_w$=32, $n_l$=1 | 5 | MAE + L1 reg. epoch=500, batch_size=64, initial_lr=1e-3 |

When computing relative coordinates given a graph $G$, the reference point $x_{ref(G)}$ is chosen to be $(x_{median}, y_{median}, 0)$ which is the median node points in $x$ and $y$ axes. This is because the geometry is fixed at the printing direction which is along $z$ axis. In addition to the relative coordinate, we encode its L1 norm $\|x_i - x_{ref(G)}\|_1$ into the node feature $v_i$. A node $i$ may belong to multiple cells which are from a different cell type. We further encode cell types into $v_i$ using one hot vectors $I_{cell\_type_i}$. The $v_i$ also includes the number of connecting nodes, $n_{e_i}$. The additional information may help the GNN to better identify abstract features across different graphs. The input features and target output are normalized to an appropriate range for GNN training.

*2) Airfoil design in aerodynamic*

The airfoil design is an important problem in aerodynamic [5]. Variables of interest of the design problem include the pressure distribution on the airfoil surface, the total lift and drag coefficients. As an airfoil is a thin object with a sharp and long tail, precise representation of its shape is crucial for accurate prediction of its aerodynamic performance. In this work, we train the GNNs to predict the surface pressure on airfoil, as well as 2 derived performance indicators: drag and lift coefficients $C_D$, $C_L$. We leverage the huge database of Reynolds-averaged Navier-Stokes (RANS) simulations generated by [3], comprising of 1,505 2D airfoil shapes obtained from the UIUC Airfoil Coordinates Database [17]. Each airfoil shape was simulated with several freestream conditions $(u_0, v_0)$. We extracted and post-processed a total of 6,372 simulation solutions (train set) with 1,448 airfoils for model training. Specifically, we construct a graph for the airfoil shape by using the surface node points. Each node is only connected to its nearest left and right neighbor nodes along the airfoil surface (Figure 1b). We also extracted and post-processed 90 simulation solutions (test set 1) with 30 airfoils that are not used for training, to evaluate the generalization of the model.

When building the graph $G$ for a given airfoil, we use $x_{ref(G)} = (0,0)$ as reference point. This is because all the airfoil shapes are already scaled to $x \in [0,1]$, with the leading edge positioned at (0,0). We also encode the given freestream condition $(u_0, v_0)$ into the node features $v_i$, for all the nodes $i \in V^G$. In addition, we use one hot vectors $I_{u/l_i}$ to indicate whether a given node $i$ belongs to the upper or lower surface. The input features are normalized to an appropriate range for GNN training. Like [3], we normalize the target pressure by dividing it with the freestream velocity $vel = u_0^2 + v_0^2$, followed by a de-mean. We do not apply normalization to the target $C_D$ and $C_L$.

*B. Model architecture, training, and evaluation*

We build a node level target model for residual stress from the feature design dataset, and a node level target model for pressure from the airfoil design dataset. For the same dataset, we build separate graph level target model for $C_D$ and $C_L$. The architecture and training configurations used for these 4 models are summarized in Table I. For each model, we tested several architecture variants and training parameters. It was found that they are robust to many choices, such as the depth $n_d$ and width $n_w$ of each MLP, the size of latent features $n_l$, the number of message passing steps $L$, and the batch size, as long as they are within a certain range.

All the MLPs use *sine* activation, which is found to be a better choice compared to other activations such as *tanh* and ReLU, in terms of solution quality. However, a *linear* activation is used at the final output layer for pressure, $C_D$ and $C_L$ models since their values span from negative to positive. The residual stress is strictly nonnegative hence we use a ReLU activation at the final output layer of the model.

We initialize the models using *He* method and train them with ADAM optimizer for a fixed number of epochs. We use the mean absolute error (MAE) loss with L1 regularization, which tends to give better training and generalization outcomes. The models are sensitive to the learning rate. We set an initial learning rate at 5e-4 or 1e-3, then reducing it by half on plateauing. The training batch size is set to be 16, 32, or 64. During the mini-batch training, all samples (graphs) within the batch are merged into a single disconnected global graph for performing forward and backward passes.

TABLE II: Performance of the trained models evaluated by the relative L2 norm error, $E$.

| Problem domain | Target output | Output type | Train set: no. sample (graph) | Train set: $E$ median (min., max.) | Test set 1: no. sample (graph) | Test set 1: $E$ median (min., max.) | Test set 2: no. sample (graph) | Test set 2: $E$ median (min., max.) |
|---|---|---|---|---|---|---|---|---|
| Feature design (3D) | Residual stress | Node level | 458 ↑ 46 – 3492 nodes | 9.13% (3.9, 31.9) | 12 ↑ 883 – 1862 nodes | 8.71% (5.9, 11.8) | 7 ↑ 1213 – 7247 nodes | 28.49% (14.6, 49.0) |
| Airfoil design (2D) | Pressure | Node level | 6372 ↑ 194 – 660 nodes | – | 90 ↑ 512 – 565 nodes | 8.71% (3.5, 28.0) | – | – |
| Airfoil design (2D) | Drag coefficients, $C_D$ | Graph level | 6372 | 9.43% | 90 | 14.23% | – | – |
| Airfoil design (2D) | Lift coefficients, $C_L$ | Graph level | 6372 | 6.84% | 90 | 8.41% | – | – |

Finally, the trained models are evaluated by the relative L2 norm error,

$$\epsilon_R = \frac{\|y_{target} - y_{predict}\|_2}{\|y_{target}\|_2} \times 100\%. \quad (5)$$

For the node level target models, we first compute their $\epsilon_R$ based on all node-wise predictions in each individual graph. We then report the statistics, i.e., median (min., max.), of all the graphs' $\epsilon_R$. For the graph level target models, we report a single $\epsilon_R$ value computed from all the predictions. The performance of models on respective data sets are reported in Table II.

### C. Modelling results on feature design problem

The residual stress model achieves a performance of about 9% median $\epsilon_R$ on both train and test set 1. From the visualization results presented in Figure 3a, we observe a good agreement between the model prediction and the ground truth solution for the geometries in test set 1. This shows that the GNN-based surrogate model can adequately learn from the training geometries and predict the quantitative outcome on the test geometries with a similar accuracy. Although not previously seen during the training process, the geometries in test set 1 and train are homogenous, since they are designed from the same set of basic features with different sizes, orientations, and intersections.

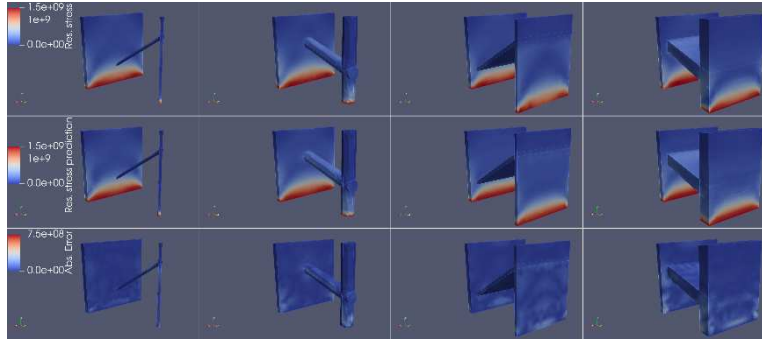

(a) Selected feature designs from Test set 1, $\epsilon_R$ = 5.90, 8.71, 8.71, 11.81% (left to right)

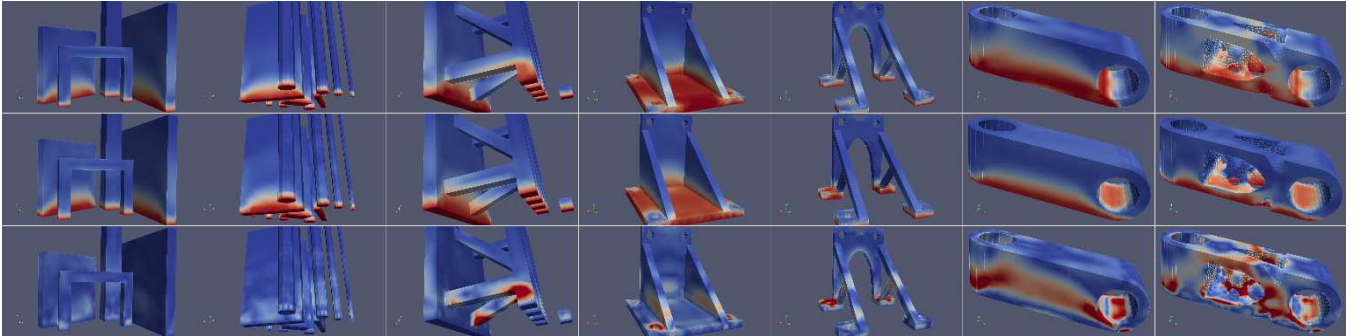

(b) Feature designs from test set 2, $\epsilon_R$ = 14.55, 20.07, 28.49, 17.14, 34.82, 43.12, 48.96% (left to right)

Figure 3: For both (a) & (b), the simulation solutions used as ground truth are displayed at the top rows. The model predictions and their absolute errors are displayed at the middle and bottom rows, respectively.

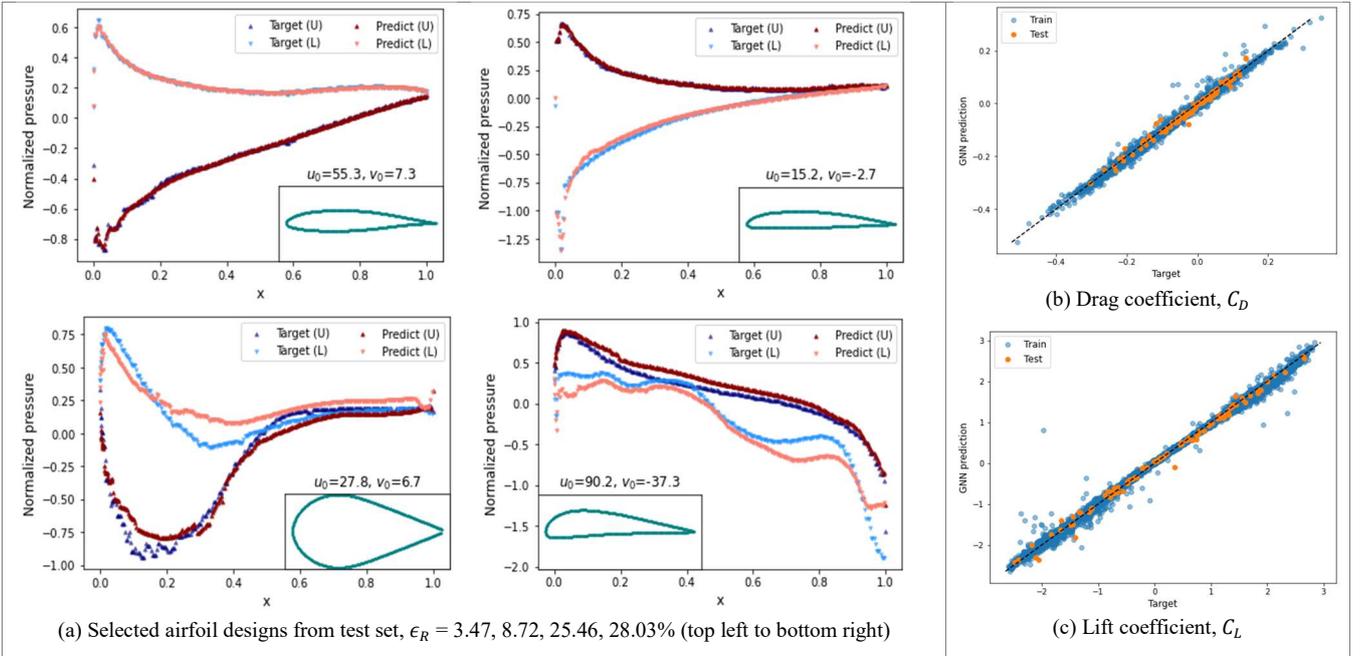

(a) Selected airfoil designs from test set, $\epsilon_R$ = 3.47, 8.72, 25.46, 28.03% (top left to bottom right)

(b) Drag coefficient, $C_D$

(c) Lift coefficient, $C_L$

Figure 4: (a) The pressure prediction on upper (U) and lower (L) surfaces of the airfoil is plotted against the simulation ground truth. The nested plot shows the airfoil design and the freestream condition that are used as the model input. (b-c) The model prediction for $C_D$ and $C_L$ are plotted against the simulation ground truth. The diagonal dotted line indicates a perfect fit to the simulation ground truth.

The prediction results of more challenging feature designs in test set 2 are visualized in Figure 3b. For the 3 geometries comprising more complex combinations of the basic features, their $\epsilon_R$'s are 14.55%, 20.07%, and 28.49%—almost 2-3 times higher than the median $\epsilon_R$ seen in train and test set 1. Note that the model has difficulty predicting the high residual stress around the intersection and joint areas, which could have great impact on the quality of product—the designer usually needs to pay attention to the high stress area. The original and optimized engineering bracket designs contain new features, such as triangle, hole, and arch, that are unseen by the model during the training. Their $\epsilon_R$'s are 17.14% and 34.82% respectively. Large prediction errors are found near the bottom holes. Moreover, the $\epsilon_R$ for original and optimized bike stem designs are 43.12% and 48.96%, respectively. These 2 designs are bulky, and they contain a lot of curvatures, which are significantly different from the training features in both size and shape.

Overall, the model performance is worse when being applied to out-of-distribution designs. Although it is possible to further reduce the prediction error on train and test set 1 by using a different set of model and training hyperparameters, these models usually do not demonstrate any accompanying performance improvement on test set 2. The performance of our GNN models is consistent to the prior study on a similar dataset while using the state-of-the-art U-Net model [7]. The U-Net uses a different (voxel) representation and is trained on a slightly larger set of geometries with data augmentation. The model gives prediction error between 14-29% on 4 out-of-distribution designs (which is the 1st, 3rd-5th geometry in our test set 2).

Although it is difficult to directly compare the computation time for different methods due to differences in implementation, our GNN model leads to a significant reduction in runtime as compared to the full order simulation, which requires approximately 1-4hours for simulating the residual stress distribution for each geometry [7], as compared to the instant prediction (< 1second) by the GNN model.

### D. Modelling results on airfoil design problem

The pressure model achieves a performance within the range of $\epsilon_R$=3.47% and 28.03% on the test set, which consists of 30 airfoil designs that are not seen by the model during the training, and there are 3 random freestream conditions for each design. Their median $\epsilon_R$ is 8.71%. Figure 4a compares the prediction with the simulation ground truth, for 4 different airfoil designs and freestream conditions. The predicted pressure values are very accurate judging by the visual inspection, even though the quantitative $\epsilon_R$ is around 9%. Note that the high-pressure profile may occur at either the upper or lower surface, depending on the freestream condition $v_o$. Figure 4b-c present the prediction results for $C_D$ and $C_L$. Their $\epsilon_R$'s are 14.23% and 8.41% respectively on the same test set.

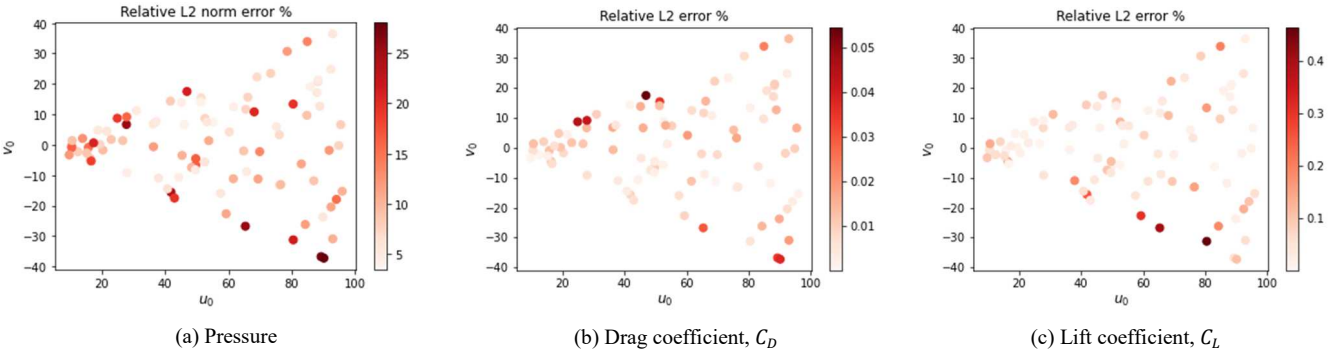

(a) Pressure

(b) Drag coefficient, $C_D$

(c) Lift coefficient, $C_L$

Figure 5: The relative L2 norm error, $\epsilon_R$ for pressure, $C_D$ and $C_L$ between model prediction and simulation ground truth on test data. Predictions with high $\epsilon_R$, as indicated by reds and dark reds, are more commonly found near to the edge of the distribution in the 2D freestream condition input space.

It is observed that there are some designs, e.g., a very thick airfoil, and certain freestream conditions where the model prediction has a large $\epsilon_R$. These samples generally have a more complex pressure profile. Even so, the pressure model can still capture the general trend, as shown in the bottom panel of Figure 4a. Although it is difficult to verify whether those with large $\epsilon_R$ are indeed entirely due to out-of-distribution designs, we note that the worse predictions are usually clustered around the edge of sample distribution in the freestream condition space, as shown in Figure 5.

Overall, we obtain satisfying results on surface pressure, $C_D$ and $C_L$ models. We believe that the issue of out-of-distribution designs will not be too severe, given that these models are trained on a large set of airfoil designs and freestream conditions. These models have the potential to replace simulations for the design of new airfoil on a range of operating conditions. The OpenFOAM simulation time for computing the flow solution on a new airfoil in current dataset takes around 40-70seconds.

## V. Conclusion

In this work, we apply GNN model for the surrogate modelling task of predicting the functional performance of different geometrical designs in engineering applications. The GNN allows us to directly train the model on unstructured mesh simulations using graph representation, which can more precisely describe the change in geometry with complex features and predict the consequences.

We present the modelling results on 2 applications: feature design in the domain of additive engineering and airfoil design in the domain of aerodynamics. For the feature design task, the model is designed to predict the residual stress—which has great impact on the quality of the printed product—for features with different shape, sizes, orientations, and intersections. For the airfoil design task, we develop models for predicting the pressure on airfoil surface, as well as key performance indicators in drag and lift coefficients. These models have been shown to provide good accuracy on a separate set of test geometries which share similar features and characteristics to the training geometries. However, we have found that generalization remains a key issue for the residual stress model where the number of training geometries is smaller. The prediction error on out-of-distribution designs is almost 2-5 times higher than the geometries used for training. Further tests on different scenarios and application domains, ideally with an extended dataset, to confirm the potential of GNN-based surrogate model can be beneficial.

The strength of the surrogate models is their prediction speed. They can make predictions for new geometry designs almost instantly, as compared to O(hour) for the high-fidelity simulations typically used in most engineering applications. Our future work includes integrating the GNN-based surrogate models into generative design process in conjunction with other CI techniques such as generative adversarial network.

## Acknowledgment

This research is supported by A*STAR under the AME Programmatic programme: "Explainable Physics-based AI for Engineering Modelling and Design (ePAI)" Award No. A20H5b0142 and the IAF-PP project: "Industrial Digital Design and Additive Manufacturing Workflows (IDDAMW)" Award No. A19E10097.